\documentclass[a4paper]{article}

\usepackage{hyperref}
\usepackage{url}

\usepackage{siunitx}
\usepackage{amsmath}
\usepackage{xcolor}
\usepackage{graphicx}
\usepackage[export]{adjustbox}
\usepackage{graphbox} 

\usepackage[utf8]{inputenc} 
\usepackage[T1]{fontenc}    
\usepackage{booktabs}       
\usepackage{amsfonts}       
\usepackage{nicefrac}       
\usepackage{microtype}      

\usepackage{amssymb}
\usepackage{amsthm}
\usepackage{bbold}
\usepackage{graphicx}
\graphicspath{ {Figures/} }
\usepackage[export]{adjustbox}

\DeclareMathOperator*{\argmin}{arg\,min}
\DeclareMathOperator*{\argmax}{arg\,max}

\newcommand{\jan}[1]{\textcolor{orange}{Jan: #1}}
\renewcommand{\jan}[1]{ }

\usepackage{INTERSPEECH_v2}

\title{Towards better decoding and language model integration in sequence to sequence models}
\name{Jan Chorowski, Navdeep Jaitly}
\address{
      Google Brain\\
      Google Inc.\\
      Mountain View, CA 94043, USA}
\email{jan.chorowski@cs.uni.wroc.pl,ndjaitly@google.com}

\begin{document}

\maketitle
\begin{abstract}
  The recently proposed Sequence-to-Sequence (seq2seq) framework advocates replacing complex data processing pipelines, such as an entire automatic speech recognition system, with a single neural network trained in an end-to-end fashion. In this contribution, we analyse an attention-based seq2seq speech recognition system that directly transcribes recordings into characters. We observe two shortcomings: overconfidence in its predictions and a tendency to produce incomplete transcriptions when language models are used. We propose practical solutions to both problems achieving competitive speaker independent word error rates on the Wall Street Journal dataset: without separate language models we reach 10.6\% WER, while together with a trigram language model, we reach 6.7\% WER. 
\end{abstract}


\section{Introduction}

Deep learning \cite{lecun2015deep} has led to many breakthroughs including speech and image recognition \cite{hinton2012deep,graves2014towards,hannun2014deep,amodei2015deep,miao2015eesen,krizhevsky2012imagenet}. A subfamily of deep models, the Sequence-to-Sequence (seq2seq) neural networks have proved to be very successful on complex transduction tasks, such as machine translation \cite{sutskever2014sequence,bahdanau2014neural,wu_google_2016}, speech recognition \cite{chorowski2015attention,bahdanau2016end,chan2015listen}, and lip-reading \cite{chung2016lip}. Seq2seq networks can typically be decomposed into modules that implement stages of a data processing pipeline: an encoding module that transforms its inputs into a hidden representation, a decoding (spelling) module which emits target sequences and an attention module that computes a soft alignment between the hidden representation and the targets. Training directly maximizes the probability of observing desired outputs conditioned on the inputs. This discriminative training mode is fundamentally different from the generative "noisy channel" formulation used to build classical state-of-the art speech recognition systems. As such, it has benefits and limitations that are different from classical ASR systems. 

Understanding and preventing limitations specific to seq2seq models is crucial for their successful development. Discriminative training allows seq2seq models to focus on the most informative features. However, it also increases the risk of overfitting to those few distinguishing characteristics. We have observed that seq2seq models often yield very sharp predictions, and only a few hypotheses need to be considered to find the most likely transcription of a given utterance. However, high confidence reduces the diversity of transcripts obtained using beam search.

During typical training the models are conditioned on ground truth transcripts and are scored on one-step ahead predictions. By itself, this training criterion does not ensure that all relevant fragments of the input utterance are transcribed. Subsequently, mistakes that are introduced during decoding may cause the model to skip some words and jump to another place in the recording. The problem of incomplete transcripts is especially apparent when external language models are used.

\section{Model Description}

Our speech recognition system, builds on the recently proposed Listen, Attend and Spell network \cite{chan2015listen}. It is an attention-based seq2seq model that is able to directly transcribe an audio recording $\boldsymbol{x}$ into a space-delimited sequence of characters $\boldsymbol{y}$. Similarly to other seq2seq neural networks, it uses an encoder-decoder architecture composed of three parts: a \emph{listener} module tasked with acoustic modeling, a \emph{speller} module tasked with emitting characters and an \emph{attention} module serving as the intermediary between the speller and the listener:
\begin{align}
\boldsymbol{h} &= \textrm{Listen}(\boldsymbol{x}) \\ 
p(\boldsymbol{y}|\boldsymbol{x}) &= \textrm{AttendAndSpell}(\boldsymbol{y}, \boldsymbol{h})
\end{align}

\subsection{The Listener}
The listener is a multilayer Bi-LSTM network that transforms a sequence of $N$ frames of acoustic features $\boldsymbol{x}_1,\boldsymbol{x}_2,\ldots, \boldsymbol{x}_N$ into a possibly shorter sequence of hidden activations $\boldsymbol{h}_1, \boldsymbol{h}_2, \ldots, \boldsymbol{h}_{N/k}$, where $k$ is a time reduction constant \cite{bahdanau2016end,chan2015listen}. 

\subsection{The Speller and the Attention Mechanism}

The speller computes the probability of a sequence of characters conditioned on the activations of the listener. The probability is computed one character at a time, using the chain rule:
\begin{equation}
    p(\boldsymbol{y}|\boldsymbol{h}) = \prod_i p(\boldsymbol{y}_i|\boldsymbol{y}_{<i}, \boldsymbol{h}).
\end{equation}
To emit a character the speller uses the attention mechanism to find a set of relevant activations of the listener $\boldsymbol{\alpha}$ and summarize them into a context $\boldsymbol{c}$. The history of previously emitted characters is encapsulated in a recurrent state $\boldsymbol{s}$:
\begin{align}
\boldsymbol{s}_i &= \text{RecurrentStep}(\boldsymbol{y}_{i-1}, \boldsymbol{s}_{i-1}, \boldsymbol{c}_{i-1}), \\ 
\boldsymbol{c}_i, \boldsymbol{\alpha}_{i} &= \textrm{Attend}(\boldsymbol{h}, \boldsymbol{s}_i, \boldsymbol{\alpha}_{i-1}), \\ 
p(\boldsymbol{y}_i|\boldsymbol{y}_{<i}, \boldsymbol{h}) &= \textrm{CharacterDistribution}(\boldsymbol{s}_i, \boldsymbol{c}_i).
\end{align}
We implement the recurrent step using a single LSTM layer. The attention mechanism is sensitive to the location of frames selected during the  previous step and employs the convolutional filters over the previous attention weights \cite{chorowski2015attention}. The output character distribution is computed using a SoftMax function. 

\subsection{Training Criterion}

Our speech recognizer computes the probability of a character conditioned on the partially emitted transcript and the whole utterance. It can thus be trained to minimize the cross-entropy between the ground-truth characters and model predictions. The training loss over a single utterance is
\begin{equation}\label{eq:train_loss}
\textrm{loss}(\boldsymbol{y}, \boldsymbol{x}) = -\sum_i\sum_c T(\boldsymbol{y}_i,c)\log p(\boldsymbol{y}_i|\boldsymbol{y}_{<i}, \boldsymbol{x}),
\end{equation}
where $T(\boldsymbol{y}_i,c)$ denotes the target label function. In the baseline model $T(\boldsymbol{y}_i,c)$ is the indicator $[\boldsymbol{y}_i=c]$, i.e. its value is $1$ for the correct character, and $0$ otherwise. When label smoothing is used, $T$ encodes a distribution over characters.

\subsection{Decoding: Beam Search}

Decoding new utterances amounts to finding the character sequence $\boldsymbol{y^*}$ that is most probable under the distribution computed by the network:
\begin{equation}
    \boldsymbol{y^*} = \argmax_{\boldsymbol{y}} p(\boldsymbol{y}|\boldsymbol{x}) = \argmin_{\boldsymbol{y}} -\log p(\boldsymbol{y}|\boldsymbol{x}).
\end{equation}

Due to the recurrent formulation of the speller function, the most probable transcript cannot be found exactly using the Viterbi algorithm. Instead, approximate search methods are used. Typically, best results are obtained using beam search. The search begins with the set (beam) of hypotheses containing only the empty transcript. At every step, candidate transcripts are formed by extending hypothesis in the beam by one character. The candidates are then scored using the model, and a certain number of top-scoring candidates forms the new beam. The model indicates that a transcript is considered to be finished by emitting a special EOS (end-of-sequence) token. 

\subsection{Language Model Integration}

The simplest solution to include a separate language model is to extend the beam search cost with a language modeling term \cite{bahdanau2016end,hannun2014deep,gulcehre2015using}:
\begin{equation}\label{eq:beam_search}
    \boldsymbol{y^*} = \argmin_{\boldsymbol{y}} -\log p(\boldsymbol{y}|\boldsymbol{x}) - \lambda\log p_\textrm{LM}(\boldsymbol{y}) - \gamma\text{coverage},
\end{equation}
where \text{coverage} refers to a term that promotes longer transcripts described it in detail in Section \ref{sec:coverage}.

We have identified two challenges in adding the language model. First, due to model overconfidence deviations from the best guess of the network drastically changed the term $-\log p(\boldsymbol{y}|\boldsymbol{x})$, which made balancing the terms in eq.~\eqref{eq:beam_search} difficult. Second, incomplete transcripts were produced unless a recording coverage term was added.

Equation~\eqref{eq:beam_search} is a heuristic involving the multiplication of a conditional and unconditional probabilities of the transcript $y$. We have tried to  justify it by adding an intrinsic language model suppression term $\log p(\boldsymbol{y})$ that would transform $p(\boldsymbol{y}|\boldsymbol{x})$ into $p(\boldsymbol{x}|\boldsymbol{y}) \propto p(\boldsymbol{y}|\boldsymbol{x}) / p(\boldsymbol{y})$. We have estimated the language modeling capability of the speller $p(\boldsymbol{y})$ by replacing the encoded speech with a constant, separately trained, biasing vector. The per character perplexity obtained was about 6.5 and we didn't observe consistent gains from this extension of the beam search criterion.

\section{Solutions to Seq2Seq Failure Modes}
We have analysed the impact of model confidence by separating its effects on model accuracy and beam search effectiveness. We also propose a practical solution to the partial transcriptions problem, relating to the coverage of the input utterance.

\subsection{Impact of Model Overconfidence}

Model confidence is promoted by the the cross-entropy training criterion. For the baseline network the training loss~\eqref{eq:train_loss} is minimized when the model concentrates all of its output distribution on the correct ground-truth character. This leads to very peaked probability distributions, effectively preventing the model from indicating sensible alternatives to a given character, such as its  homophones. Moreover, overconfidence can harm learning the deeper layers of the network. The derivative of the loss backpropagated through the SoftMax function to the logit corresponding to character $c$ equals $[\boldsymbol{y}_i=c]-p(\boldsymbol{y}_i|\boldsymbol{y}_{<i}, \boldsymbol{x})$, which approaches $0$ as the network's output becomes concentrated on the correct character. Therefore whenever the spelling RNN makes a good prediction, very little training signal is propagated through the attention mechanism to the listener.

Model overconfidence can have two consequences. First, next-step character predictions may have low accuracy due to overfitting. Second, overconfidence may impact the ability of beam search to find good solutions and to recover from errors. 

\begin{figure*}
    \centering
    \includegraphics[width=0.8\textwidth]{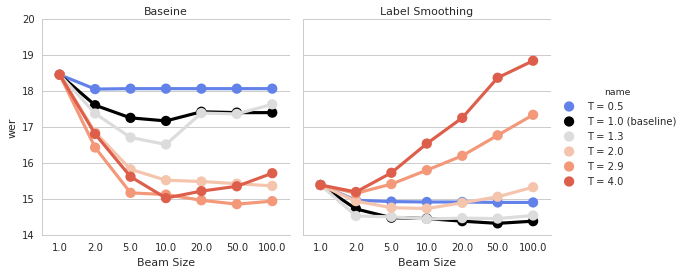}
    \caption{Influence of beam width and SoftMax temperature on decoding accuracy. In the baseline case (no label smoothing) increasing the temperature reduces the error rate. When label smoothing is used the next-character prediction improves, as witnessed by WER for beam size=1, and tuning the temperature does not bring additional benefits.}
    \label{fig:baseline_temp}
\end{figure*}

We first investigate the impact of confidence on beam search by varying the temperature of the SoftMax function. Without retraining the model, we change the character probability distribution to depend on a \emph{temperature} hyperparameter $T$:
\begin{equation}
    p(y_i) = \frac{\exp(l_i/T)}{\sum_j \exp(l_j/T)}.
\end{equation}
At increased temperatures the distribution over characters becomes more uniform. However, the preferences of the model are retained and the ordering of tokens from the most to least probable is preserved. Tuning the temperature therefore allows to demonstrate the impact of model confidence on beam search, without affecting the accuracy of next step predictions.

Decoding results of a baseline model on the WSJ dev93 data set are presented in Figure~\ref{fig:baseline_temp}. We haven't used a language model. At high temperatures deletion errors dominated. We didn't want to change the beam search cost and instead constrained the search to emit the EOS token only when its probability was within a narrow range from the most probable token. We compare the default setting ($T=1$), with a sharper distribution ($T=0.5$) and smoother distributions ($T\in\{1.3,\ldots,4\}$). All strategies lead to the same greedy decoding accuracy, because temperature changes do not affect the selection of the most probable character. As temperature increases beam search finds better solutions, however care must be taken to prevent truncated transcripts.

\subsection{Label Smoothing Prevents Overconfidence}

A elegant solution to model overconfidence was problem proposed for the Inception image recognition architecture \cite{szegedy_rethinking_2015}. For the purpose of computing the training cost the ground-truth label distribution is smoothed, with some fraction of the probability mass assigned to classes other than the correct one. This in turn prevents the model from learning to concentrate all probability mass on a single token. Additionally, the model receives more training signal because the error function cannot easily saturate.

Originally \emph{uniform} label smoothing scheme was proposed in which the model is trained to assign $\beta$ probability mass to he correct label, and spread the $1 - \beta$ probability mass uniformly over all classes \cite{szegedy_rethinking_2015}. Better results can be obtained with \emph{unigram} smoothing which distributes the remaining probability mass proportionally to the marginal probability of classes \cite{pereyra2017regularizing}. In this contribution we propose a \emph{neighborhood} smoothing scheme that uses the temporal structure of the transcripts: the remaining $1-\beta$ probability mass is assigned to tokens neighboring in the transcript. Intuitively, this smoothing scheme helps the model to recover from beam search errors: the network is more likely to make mistakes that simply skip a character of the transcript.

We have repeated the analysis of SoftMax temperature on beam search accuracy on a network trained with neighborhood smoothing in Figure~\ref{fig:baseline_temp}. We can observe two effects. First, the model is regularized and greedy decoding leads to nearly 3 percentage smaller error rate. Second, the entropy of network predictions is higher, allowing beam search to discover good solutions without the need for temperature control. Moreover, the since model is trained and evaluated with $T=1$ we didn't have to control the emission of EOS token.

\subsection{Solutions to Partial Transcripts Problem}
\label{sec:coverage}

\begin{table}[t]
\centering
\caption{Example of model failure on validation '4k0c030n'}
\label{tb:incomplete_transcripts}
\begin{tabular}{p{4.25cm}cc}
Transcript & LM cost     & Model cost      \\
           & $\log p(y)$ & $\log p(y|x)$   \\ \hline
"chase is nigeria's registrar and the society is an independent organization hired to count votes" &
-108.5 & -34.5 \\
"in the society is an independent organization hired to count votes"         & -64.6  & -19.9 \\
"chase is nigeria's registrar" & -40.6 & -31.2 \\
"chase's nature is register" & -37.8 & -20.3 \\
"" & -3.5 & -12.5 \\
\end{tabular}
\end{table}

When a language model is used wide beam searches often yield incomplete transcripts. With narrow beams, the problem is less visible due to implicit hypothesis pruning. We illustrate a failed decoding in Table~\ref{tb:incomplete_transcripts}. The ground truth (first row) is the least probable transcript according both to the network and the language model. A width 100 beam search with a trigram language model finds the second transcript, which misses the beginning of the utterance. The last rows demonstrate severely incomplete transcriptions that may be discovered when decoding is performed with even wider beam sizes.

We compare three strategies designed to prevent incomplete transcripts. The first strategy doesn't change the beam search criterion, but forbids emitting the EOS token unless its probability is within a set range of that of the most probable token. This strategy prevents truncations, but is inefficient against omissions in the middle of the transcript, such as the failure shown in Table~\ref{tb:incomplete_transcripts}. Alternatively, beam search criterion can be extended to promote long transcripts. A term depending on the \emph{transcript length} was proposed for both CTC \cite{hannun2014deep} and seq2seq \cite{bahdanau2016end} networks, but its usage was reported to be difficult because beam search was looping over parts of the recording and additional constraints were needed \cite{bahdanau2016end}. To prevent looping we propose to use a \emph{coverage} term that counts the number of frames that have received a cumulative attention greater than $\tau$:
\begin{equation}
\textrm{coverage} = \sum_j\left[\sum_i\boldsymbol{\alpha}_{ij} > \tau \right].
\end{equation}

The coverage criterion prevents looping over the utterance because once the cumulative attention bypasses the threshold $\tau$ a frame is counted as selected and subsequent selections of this frame do not reduce the decoding cost. In our implementation, the coverage is recomputed at each beam search iteration using all attention weights produced up to this step.

\begin{figure}
    \centering
    \includegraphics[width=\columnwidth]{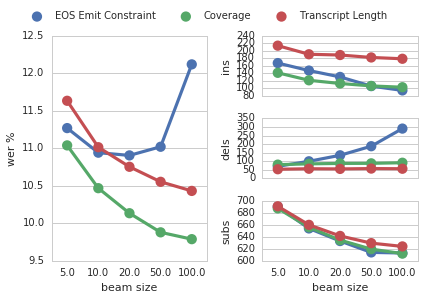}
    \caption{Impact of using techniques that prevent incomplete transcripts when a trigram language models is used on the dev93 WSJ subset. Results are averaged across two networks}
    \label{fig:coverage_criterion}
\end{figure}

In Figure~\ref{fig:coverage_criterion} we compare the effects of the three methods when decoding a network that uses label smoothing and a trigram language model. Unlike \cite{bahdanau2016end} we didn't experience looping when beam search promoted transcript length. We hypothesize that label smoothing increases the cost of correct character emissions which helps balancing all terms used by beam search. We observe that at large beam widths constraining EOS emissions is not sufficient. In contrast, both promoting coverage and transcript length yield improvements with increasing beams. However, simply maximizing transcript length yields more word insertion errors and achieves an overall worse WER. 

\section{Experiments}

We conducted all experiments on the Wall Street Journal dataset, training on si284, validating on dev93 and evaluating on eval92 set. The models were trained on 80-dimensional mel-scale filterbanks extracted every 10ms form 25ms windows, extended with their temporal first and second order differences and per-speaker mean and variance normalization. Our character set consisted of lowercase letters, the space, the apostrophe, a noise marker, and start- and end- of sequence tokens. For comparison with previously published results, experiments involving language models used an extended-vocabulary trigram language model built by the Kaldi WSJ s5 recipe \cite{povey2011kaldi}. We have use the FST framework to compose the language model with a  "spelling lexicon" \cite{miao2015eesen,bahdanau2016end,openfst}. All models were implemented using the Tensorflow framework \cite{abadi2016tensorflow}.

Our base configuration implemented the Listener using 4 bidirectional LSTM layers of 256 units per direction (512 total), interleaved with 3 time-pooling layers which resulted in an 8-fold reduction of the input sequence length, approximately equating the length of hidden activations to the number of characters in the transcript. The Speller was a single LSTM layer with 256 units. Input characters were embedded into 30 dimensions. The attention MLP used 128 hidden units, previous attention weights were accessed using 3 convolutional filters spanning 100 frames. LSTM weights were initialized uniformly over the range $\pm0.075$. Networks were trained using 8 asynchronous replica workers each employing the ADAM algorithm \cite{kingma2014adam} with default parameters and the learning rate set initially to $10^{-3}$, then reduced to $10^{-4}$ and $10^{-5}$ after 400k and 500k training steps, respectively. Static Gaussian weight noise with standard deviation 0.075 was applied to all weight matrices after 20000 training steps. We have also used a small weight decay of $10^{-6}$.

We have compared two label smoothing methods: \emph{unigram} smoothing \cite{pereyra2017regularizing} with the probability of the correct label set to $0.95$ and \emph{neighborhood} smoothing with the probability of correct token set to $0.9$ and the remaining probability mass distributed symmetrically over neighbors at distance $\pm 1$ and $\pm 2$ with a $5:2$ ratio. We have tuned the smoothing parameters with a small grid search and have found that good results can be obtained for a broad range of settings.

We have gathered results obtained without language models in Table~\ref{table:WSJresultsNoLM}. We have used a beam size of 10 and no mechanism to promote longer sequences. We report averages of two runs taken at the epoch with the lowest validation WER. Label smoothing brings a large error rate reduction, nearly matching the performance achieved with very deep and sophisticated encoders \cite{zhang2016very}.

Table~\ref{table:WSJresultsLM} gathers results that use the extended trigram language model.  We report averages of two runs. For each run we have tuned beam search parameters on the validation set and applied them on the test set. A typical setup used beam width 200, language model weight $\lambda=0.5$, coverage weight $\gamma=1.5$ and coverage threshold $\tau = 0.5$. Our best result surpasses CTC-based networks \cite{miao2015eesen} and matches the results of a DNN-HMM and CTC ensemble \cite{graves_endtoend_2014}.

\begin{table}[t]
\centering
\caption{Results without separate language model on WSJ.}
\label{table:WSJresultsNoLM}
\begin{tabular}{l|ccc}
\bf Model & \bf Parameters & \bf dev93 &  \bf eval92 \\
\midrule

CTC \cite{graves2014towards} & 26.5M & - & 27.3 \\
seq2seq \cite{bahdanau2016end} & 5.7M & - & 18.6 \\
seq2seq \cite{chan2016latent} & 5.9M & - & 12.9 \\
seq2seq \cite{zhang2016very} & - & - & 10.5 \\
\midrule
Baseline & 6.6M & 17.9 & 14.2 \\
Unigram LS & 6.6M & {\bf 13.7} & {\bf 10.6} \\
Temporal LS & 6.6M & 14.1 & 10.7 \\
\end{tabular}
\end{table}

\begin{table}[t]
\centering
\caption{Results withextended trigram language model on WSJ.}
\label{table:WSJresultsLM}
\begin{tabular}{l|cc}
\bf Model & \bf dev93 &  \bf eval92 \\
\midrule
seq2seq \cite{bahdanau2016end} & - & 9.3 \\
CTC \cite{graves2014towards} & - & 8.2 \\
CTC \cite{miao2015eesen} & - & 7.3 \\
\midrule

Baseline + Cov & 12.6 & 8.9 \\
Unigram LS + Cov. & 9.9 & 7.0 \\ 
Temporal LS + Cov. & \bf{9.7} & {\bf 6.7} \\ 

\end{tabular}
\end{table}

\section{Related Work}

Label smoothing was proposed as an efficient regularizer for the Inception architecture \cite{szegedy_rethinking_2015}. Several improved smoothing schemes were proposed, including sampling erroneous labels instead of using a fixed distribution \cite{xie2016disturblabel}, using the marginal label probabilities \cite{pereyra2017regularizing}, or using early errors of the model \cite{aghajanyan_softtarget_2016}. Smoothing techniques increase the entropy of a model's predictions, a technique that was used to promote exploration in reinforcement learning \cite{williams1991function,mnih2016asynchronous,luo2016learning}. Label smoothing prevents saturating the SoftMax nonlinearity and results in better gradient flow to lower layers of the network \cite{szegedy_rethinking_2015}. A similar concept, in which training targets were set slightly below the range of the output nonlinearity was proposed in \cite{lecun_efficient_2012}.

Our seq2seq networks are locally normalized, i.e. the speller produces a probability distribution at every step. Alternatively normalization can be performed globally on whole transcripts. In discriminative training of classical ASR systems normalization is performed over lattices \cite{he2008discriminative}. In the case of recurrent networks lattices are replaced by beam search results. Global normalization has yielded important benefits on many NLP tasks including parsing and translation \cite{andor_globally_2016, wiseman_sequence_2016}. Global normalization is expensive, because each training step requires running beam search inference. It remains to be established whether globally normalized models can be approximated by cheaper to train locally normalized models with proper regularization such as label smoothing.

Using source coverage vectors has been investigated in neural machine translation models. Past attentions vectors were used as auxiliary inputs in the emitting RNN either directly \cite{luong_effective_2015}, or as cumulative coverage information \cite{tu_coverage_2016}. Coverage embeddings vectors associated with source words end modified during training were proposed in \cite{mi_coverage_2016}. Our solution that employs a coverage penalty at decode time only is most similar to the one used by the Google Translation system \cite{wu_google_2016}.

\section{Conclusions}

We have demonstrated that with efficient regularization and careful decoding the sequence-to-sequence approach to speech recognition can be competitive with other non-HMM techniques, such as CTC. 

\section{Acknowledgements}

\bibliographystyle{IEEEtran}
\bibliography{main}

\end{document}